\documentclass[10pt,twocolumn,letterpaper]{article}

\usepackage{iccv}               % To produce the CAMERA-READY version
\usepackage{svg}
\usepackage{placeins}
\usepackage{tabularx} 
\usepackage{float}
\usepackage{mathtools}
\usepackage{colortbl}
\newcolumntype{Y}{>{\raggedright\arraybackslash}X}
\usepackage{multirow}
\usepackage{graphicx}  % Optional, if you're including figures
\usepackage{amsmath}   % Optional, for math support
\usepackage{booktabs}  % Optional, for cleaner tables
\usepackage{caption}   % Optional, for better caption formatting

\definecolor{iccvblue}{rgb}{0.21,0.49,0.74}
\usepackage[pagebackref,breaklinks,colorlinks,allcolors=iccvblue]{hyperref}

\def\paperID{13} % *** Enter the Paper ID here
\def\confName{ICCV}
\def\confYear{2025}

\title{STER-VLM: Spatio-Temporal With Enhanced Reference Vision-Language Models}

\author{
\thanks{All authors contributed equally to this paper. \\}
Tinh-Anh Nguyen-Nhu$^{1, 5*}$ \quad Triet Dao Hoang Minh$^{2*}$ \quad  Dat To-Thanh$^{3, 5*}$ \\ Phuc Le-Gia$^{3, 5}$ \quad Tuan Vo-Lan$^{3, 5}$ \quad Tien-Huy Nguyen$^{4, 5\dagger}$ \\ \\
$^{1}$ Ho Chi Minh University of Technology, Vietnam \\
$^{2}$ Vietnamese-German University, Vietnam \\
$^{3}$ Ho Chi Minh University of Science,  Vietnam \\
$^{4}$ University of Information Technology, Vietnam \\
$^{5}$Vietnam National University, Ho Chi Minh city, Vietnam \\
$^{\dagger}$ Corresponding author}
\begin{document}
\setlength{\parskip}{3pt}
\maketitle

\begin{abstract}
Vision-language models (VLMs) have emerged as powerful tools for enabling automated traffic analysis; however, current approaches often demand substantial computational resources and struggle with fine-grained spatio-temporal understanding. This paper introduces STER-VLM, a computationally efficient framework that enhances VLM performance through (1) caption decomposition to tackle spatial and temporal information separately, (2) temporal frame selection with best-view filtering for sufficient temporal information, and (3) reference-driven understanding for capturing fine-grained motion and dynamic context and (4) curated visual/textual prompt techniques. Experimental results on the WTS \cite{kong2024wts} and BDD \cite{BDD} datasets demonstrate substantial gains in semantic richness and traffic scene interpretation. Our framework is validated through a decent test score of 55.655 in the AI City Challenge 2025 Track 2, showing its effectiveness in advancing resource-efficient and accurate traffic analysis for real-world applications.
\end{abstract}    
\section{Introduction}

With the rapid advancement of large-scale AI models, vision-language models (VLMs),  have emerged as powerful systems that integrate visual perception with language understanding to support tasks like image captioning and visual question answering \cite{LLaVA,Qwen-VL,nguyen2024improvinggeneralizationvisualreasoning}, multimodal retrieval \cite{10.1145/3701716.3717653, Nguyen-Nhu_2025_CVPR, Tran_2025_CVPR}. Foundational models such as GPT-4V~\cite{2023GPT4VisionSC}, LLaVA~\cite{LLaVA}, and Qwen-VL~\cite{Qwen-VL} demonstrate strong zero-shot capabilities by combining cutting-edge vision and language components. Recently, VLMs have gained attention in traffic safety analysis, offering a unified framework for accident detection, behavior analysis, and detailed scene interpretation—contributing to more intelligent and safer transportation systems.

Recent work on VLMs in the traffic domain ~\cite{Shuo24AIC24, CityLLaVA, xuan2024divide, zhou2025tumtraffic} builds on foundational models while addressing the unique challenges of urban and driving scenarios. For example, Divide and Conquer Boosting~\cite{xuan2024divide} proposes decomposing the video captioning task into multiple subtasks, to improve the accuracy and interpretability of the final output. Although this modular approach improves fine-grained understanding, it often requires training and deployment of several large-scale models, which can lead to considerable computational complexity and resource demands. Inspired by these benefits, yet keenly aware of its computational drawbacks, this paper proposes a novel approach that not only enhances VLM performance but also utilizes fewer computational resources. Overall, the contributions presented in this paper are as follows:
\begin{itemize}
    \item \textbf{Caption Decomposition and Training Strategy}: We propose a caption decomposition approach that disentangles spatial and temporal information to better align with structured annotations. Additionally, we design a tailored training strategy that effectively leverages this decomposed representation for improved learning.
    \item \textbf{Temporal Frame Selection and Best-view Filtering}: We introduce a novel frame selection and best-view filtering method to ensure efficient and informative visual inputs for captioning.
    \item \textbf{Reference-driven Enhancement}: We extract frame-level references to support captioning tasks, thereby ensuring the capture of fine-grained visual details.
    \item \textbf{Instruction Optimizing}: We design visual and textual prompting techniques to instruct and guide the model’s attention and improve caption quality.
    \item \textbf{Comprehensive Evaluation}: We conduct extensive experiments to validate the effectiveness of our proposed method, demonstrating consistent improvements in captioning, visual question answering, and overall visual understanding.
\end{itemize}

% To support these emerging applications, high-quality datasets with rich multi-modal annotations are essential. The Woven Traffic Safety (WTS) dataset~\cite{kong2024wts} addresses this need by focusing on pedestrian-centric traffic scenes with fine-grained annotations. It provides multi-view video recordings along with detailed textual descriptions that capture both spatial and temporal aspects of pedestrian and vehicle behaviors. WTS is a valuable benchmark for video captioning, accident analysis, and multimodal learning in diverse traffic scenes.

\section{Related Work}

\subsection{Traffic Video Dataset For Analysis}

Recent benchmarks address limitations in understanding traffic video, including regional bias, weak temporal modeling, and limited reasoning capabilities. RoadSocial~\cite{parikh2025roadsocial} introduces diverse user-generated videos for hallucination-resistant question-resolution (QA), while SurveillanceVQA-589K~\cite{liu2025surveillancevqa} emphasizes behavioral semantics through interactive QA. TUMTraffic-VideoQA~\cite{zhou2025tumtraffic} enhances spatio-temporal grounding and object reasoning, and WTS~\cite{kong2024wts} captures fine-grained pedestrian behavior via multi-view, 3D gaze, and detailed captions. MAPLM~\cite{cao2024maplm} enables map-aware reasoning with panoramic images, LiDAR, and HD maps. SUTD-TrafficQA ~\cite{Xu_2021_CVPR} establishes one of the earliest large-scale video-based QA benchmarks in traffic scenarios. It consists of 10,080 in-the-wild videos and 62,535 QA pairs desgined around six challenging reasoning tasks. This datasets emphasizes both causal reasoning and computational efficiency via the proposed Eclipse model, making it particularly relevant for intelligent transportation and assisted driving applications where real-time constraints are critical. In addition to video-centric resources, KDD 2023’s situational reasoning benchmark~\cite{zhang2023study} explore traffic understanding through a text-based perspective. The authors introduce novels dataset that focus on decision-making, causal reasoning, and driving test problem-solving.

% SUTD-TrafficQA~\cite{xu2021sutd} focuses on causal and counterfactual reasoning with a glimpse-based model that enhances computational efficiency while maintaining high-level understanding.

\subsection{Vision Large Language Model in Traffic Safety Analysis}

Vision-language models (VLMs) have made notable progress in traffic safety but still face challenges in spatial-temporal reasoning, limited data, and modality alignment. CityLLaVA~\cite{duan2024cityllava} uses domain-specific tuning with visual/text prompts and block expansion. Divide and Conquer~\cite{xuan2024divide} segments captions for two-stage training. LLaVA-ST~\cite{li2025llava} introduces LAPE and STP for better alignment. SeeUnsafe~\cite{zhang2025language} aggregates severity-aware clips for accident interpretation. AccidentGPT~\cite{wang2024accidentgpt} fuses V2X and BEV for holistic prediction. CrashSage~\cite{zhen2025crashsage} transforms crash reports into narratives for severity forecasting with explanations. ScVLM~\cite{shi2025scvlmenhancingvisionlanguagemodel} adopts a hybrid approach that combines supervised learning, contrastive learning, and language modeling to improve the understanding, classification, and narration of safety-critical driving events. SafePLUG~\cite{sheng2025safeplugempoweringmultimodalllms} addresses the limitations of coarse-grained VLM reasoning by equipping MLLMs with pixel-level segmentation and temporal event localization for fine-grained traffic accident analysis.

\section{Method} 
\begin{figure*}[htb]
     \centering
     \includegraphics[width=1.0\textwidth]{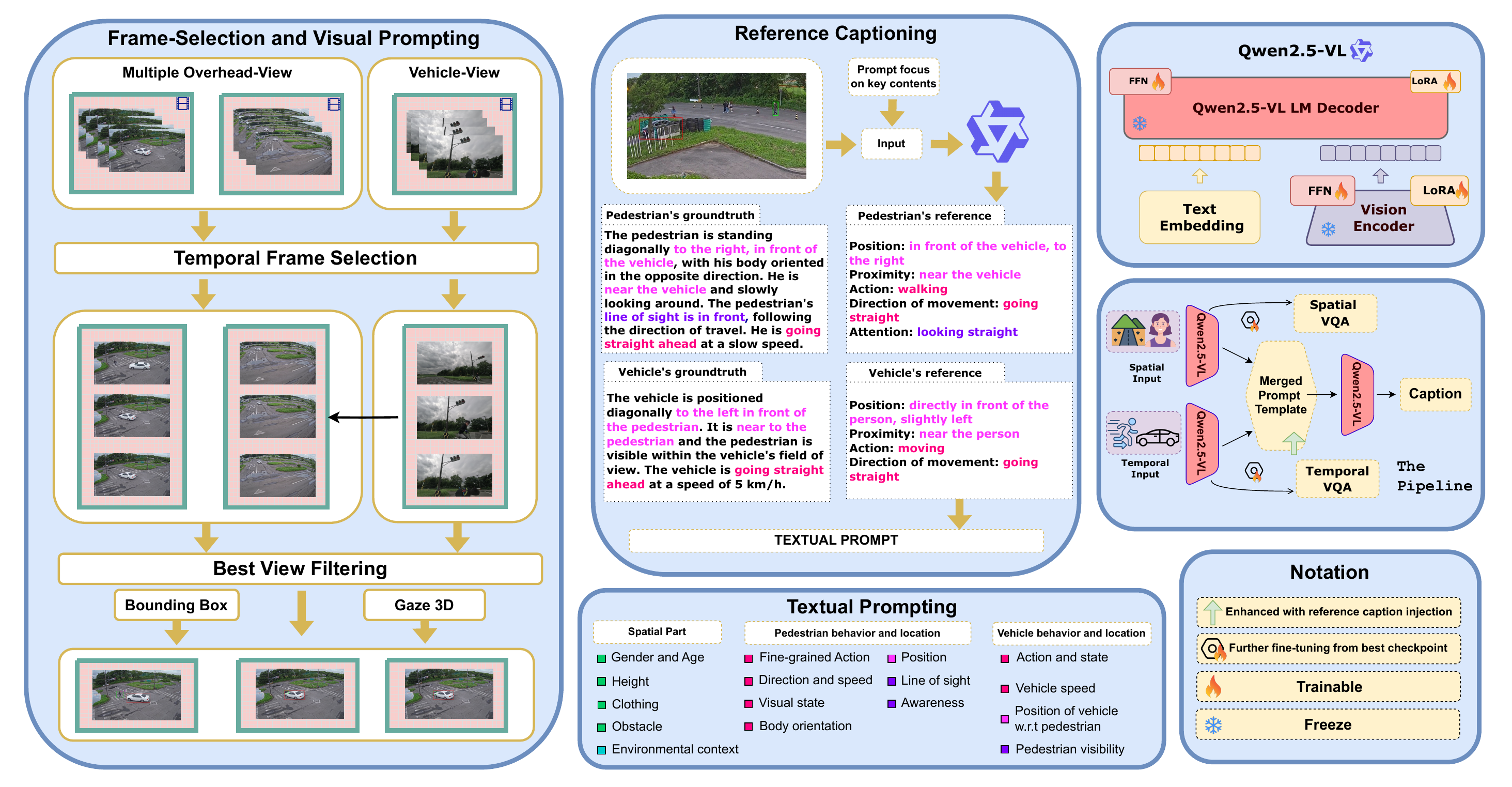} 
     \caption{The overview of the our pipeline. We employ Qwen2.5-7B-Instruct~\cite{bai2025qwen25vltechnicalreport}, enhanced with LoRA~\cite{hu2022lora}, and incorporate a combination of frame selection and filtering, textual-visual prompting and a caption decomposition training strategy.}
     \label{fig:intro} 
\end{figure*}

 % \begin{figure*}[htb]
 %     \centering
 %     \includegraphics[width=0.9\textwidth]{iccv/images/pipeline_1.pdf} 
 %     \caption{The overview of the our pipeline. We employ Qwen2.5-7B-Instruct \cite{bai2025qwen25vltechnicalreport}, enhanced with LoRA \cite{hu2022lora}, and incorporate a combination of frame selection and filtering, textual-visual prompting and a caption decomposition training strategy.}
 %     \label{fig:intro} 
 % \end{figure*}

%\begin{figure*}[t]
%    \centering
%    \includesvg[scale=0.2]{iccv/images/pipeline.drawio_compressed}
%    \caption{The overview of the our pipeline. We employ Qwen2.5-7B-Instruct \cite{bai2025qwen25vltechnicalreport}, enhanced with LoRA \cite{hu2022lora}, and incorporate a combination of frame selection and filtering, textual-visual prompting, and a caption decomposition training strategy.}
%    \label{fig:svg_image}
%\end{figure*}

This section presents our comprehensive approach for robust spatio-temporal understanding in traffic scenarios, with the overall pipeline illustrated in \textbf{Figure~\ref{fig:intro}}. With the complexity observed, our methodology consists of four key components: (1) Caption Spatial-Temporal Decomposition that provides cues for our training strategy, (2) a temporal frame selection and filtering mechanism to focus on critical frames, (3) a reference-driven enhancement that provides more cues and insights during tuning, (4) visual and textual prompt enhancement to improve context-sensitive understanding, and (5) a novel training strategy that leverages the decomposed spatio-temporal cues for optimal learning.

\subsection{Caption Decomposition}
To exploit the compositional structure of traffic video descriptions, we introduce a preprocessing phase that decomposes the ground-truth captions into semantically disjoint spatial-invariant and temporal-variant components. Using Qwen2.5-72B-Instruct~\cite{bai2025qwen25vltechnicalreport} as a caption parser, we parse each caption into two focused segments: one that responds to spatial-invariant cues and the other that focuses on temporal dynamics. This decomposition aligns with the WTS annotation format and ensures that each subtask can be modeled independently. 

For spatial-invariant extraction, we input captions from all temporal, and since spatial-invariant characteristics remain consistent across phases, these fixed elements create recognizable patterns that enable the MLLM to effectively distinguish between spatial-invariant and temporal-variant components, reducing flaws during fragmentation. We construct few-shot prompts to explicitly instruct the LLM to retain only the information pertinent to the target segment. As depicted in \textbf{Figure~\ref{fig:decomposition_caption}}, this prompting strategy offers clear and structured guidance, enabling precise and segment-specific extraction by the LLM. \textbf{Figure~\ref{fig:decomposition_caption}} shows that our approach cleanly separates original captions without confusion artifacts such as caption mixing or semantic drift, while preserving the key characteristics of both components.

\definecolor{myblue}{RGB}{218, 232, 252}
\definecolor{mygreen}{RGB}{0, 204, 102}
\definecolor{mymagenta}{RGB}{255, 0, 128}
\begin{figure*}[!t]
\centering
\small
\setlength{\tabcolsep}{0pt} % Adjust column spacing
\begin{tabularx}{\textwidth}{|>{\raggedright\arraybackslash}X|}
\hline
\rowcolor{black}
\textcolor{white}{\textbf{Original Caption}} \\
\hline
A \textcolor{mygreen}{male pedestrian} in his \textcolor{mygreen}{30s, approximately 170 cm tall}, was observed in a \textcolor{mymagenta}{squatting position on a residential road. He was diagonally positioned to the left in front of a vehicle, which was traveling in the opposite direction. The pedestrian appeared to be unaware of the vehicle, with his line of sight focused on the road surface. The distance between the pedestrian and the vehicle was far}. The pedestrian was dressed in a \textcolor{mygreen}{white jacket, white hat, and black slacks}. The \textcolor{mygreen}{weather was clear and the brightness of the surroundings was bright}. The \textcolor{mygreen}{road surface was dry and level, made of asphalt}. The \textcolor{mygreen}{traffic volume was light, as it was a two-way traffic on the road}. \textcolor{mygreen}{There were street lights present, but neither both sides of the road had a sidewalk nor a roadside strip}. Additionally, an \textcolor{mymagenta}{obstacle measuring 1 meter in height and 0.5 meters in width was positioned behind the pedestrian.}\\
\hline
\end{tabularx}

\vspace{-0.5pt} % reduce gap between tables

\begin{tabularx}{\textwidth}{|>{\raggedright\arraybackslash}X|>{\raggedright\arraybackslash}X|}
\hline
\rowcolor{black}
\textcolor{mygreen}{\textbf{Spatial-invariant}} &
\textcolor{mymagenta}{\textbf{Temporal-variant}} \\
\hline
A male pedestrian in his 30s, approximately 170 cm tall. He was dressed in a white jacket, white hat, and black slacks. The weather was clear and the brightness of the surroundings was bright. The road surface was dry and level, made of asphalt. The traffic volume was light, as it was a two-way traffic on the road. There were street lights present, but neither both sides of the road had a sidewalk nor a roadside strip. &
He was observed in a squatting position on a residential road. He was diagonally positioned to the left in front of a vehicle, which was traveling in the opposite direction. The pedestrian appeared to be unaware of the vehicle, with his line of sight focused on the road surface. The distance between the pedestrian and the vehicle was far. An obstacle measuring 1 meter in height and 0.5 meters in width was positioned behind the pedestrian. \\
\hline
\end{tabularx}

\caption{An illustration of our caption decomposition strategy. Red highlights spatial-invariant details (e.g., environment, object attributes), while blue indicates temporal-variant cues (e.g., actions, positions).}
\label{fig:decomposition_caption}
\end{figure*}

%\subsection{Temporal frame filtering and Reference} % remember to change name later
\subsection{Temporal Frame Selection and Filtering}
Although recent Vision-Language Models~\cite{bai2025qwen25vltechnicalreport, LLaVA, zhu2025internvl3exploringadvancedtraining, CogVLM} are capable of handling multiple frames, for the captioning task, several challenges still exist. Those challenges include excessively brief captions, memory overflows, and significant temporal discrepancies when too many frames are used. In contrast, if too few frames are provided, the captions may fail to capture the temporal-variant changes. Additionally, in multi-camera datasets such as WTS~\cite{kong2024wts} and BDD~\cite{BDD}, deciding which camera views to select becomes essential. In our proposed method, we want to propose 2 contributions to select frame: (1) Temporal frame selection, (2) Best-view filtering. 

\noindent \textbf{Temporal frame selection.} Although video phase segments vary in duration from 0.5 to 9 seconds, the average length is approximately 1 second for the first three phases and 2-3 seconds for the final two. To balance between representational richness and computational efficiency, we uniformly sample 3 frames from each phase in each video in the scenario. This approach mitigates the risk of losing temporal information when choosing only the first frame like prior works~\cite{CityLLaVA, xuan2024divide} and the computationally expensive training and inference stage when selecting more than 3 frames.

\noindent \textbf{Best-view filtering.} Given the limited number of frames that can be processed, we adopt a best-view filtering method to ensure that the selected frames contain sufficient and relevant visual information for accurate captioning. For spatial-invariant captioning, we select the first frame of phase 1 and 2 to capture environmental contexts, and we select the frames with the largest pedestrian's bounding box in phase 3 and 4 for pedestrian appearance. For temporal-variant captioning, a phase-specific frame selection strategy is employed. Specifically, let $\mathbf{I}_{\text{phase}}$ denote the set of selected frames for a given phase:
\begin{equation*}
\mathbf{I}_{\text{phase}} =
\begin{dcases}
    [ \mathbf{I}_{best},\ \mathbf{I}_{addition} ] & \text{if phase} \in [1, 2, 3] \\
    [ \mathbf{I}_{1},\ \mathbf{I}_{2},\ \mathbf{I}_{3} ] & \text{if phase} \in [4, 5]
\end{dcases} 
\end{equation*}
where $\mathbf{I}_{best}$ is an image with the largest pedestrian's bounding boxes in all of the scenario's videos, $\mathbf{I}_{addition}$ is an additional image with the largest sum of pedestrian's and vehicle's bounding boxes in all of the scenario's videos. For phases 4 and 5, which contain rich temporal information, we uniformly sample frames $\mathbf{I}_{1},\ \mathbf{I}_{2},\ \mathbf{I}_{3}$ in the phase for temporal information, and we only take frames in the camera with the largest sum of bounding boxes to capture the most detail of scenarios. 
% In cases where bounding box annotations are missing for a given phase, we prioritize selecting frames from vehicle's view camera views over fixed overhead cameras, as the latter may be located too far from the incident scene to provide sufficient detail. 
% to be continued....

\subsection{Reference-Driven Understanding Approach}
Recent advances in Vision-Language Models (VLMs) have significantly improved both image and video understanding. However, while these models perform well in image understanding, they limit themselves to capturing abundant video information. Although many recent VLMs excel at generating accurate overall video summaries or answering general questions about videos, they struggle with temporal dynamics and distinguishing fine-grained differences between frames. To utilize the strengths and generalization of VLMs in image understanding, they can effectively understand objects in each image and their relationships (such as distances, positions, and actions based on poses), we employed the pre-trained model Qwen2.5-VL-72B~\cite{bai2025qwen25vltechnicalreport} to generate key content for each frame, which were then used as references for our video captioning model. These references are integrated into our input prompts. They serve as hints and help models focus on the key details between each frame (temporal information) and help to capture important semantics.
% However, they were not directly used in the final captions; instead, they served only as reference points because the pre-trained model might generate inaccurate captions or understand the content differently from the ground truth.

% In our experiments, using references improved the performance of the VQA model (see figure ...) .The reason for this is that the VQA ground truth is relatively short and simple, lacking the detail needed for the model to comprehend the context fully. In this case, the reference helps compensate for this limitation, acting as a hint for the model to gain a deeper understanding of the content.

% anh doi Prompting Engineering thanh Prompt Optimization hoac Instruction Optimization, cho no fancy hon
\subsection{Instruction Optimization}
\subsubsection{Visual-Aware Prompting} 
The visual prompt serves as an effective mechanism to direct the attention of VLMs toward the most relevant parts of an image. Using cues such as the red circle dot or the colored bounding box, these prompts explicitly indicate which regions or objects need to be focused on~\cite{Yao2021CPTCP, shtedritski2023does}. This targeted attention allows the model to better understand both the appearance of pedestrians and the spatial context in the scene.

% In this work, we employ visual prompts by drawing green bounding boxes around pedestrians and red bounding boxes around vehicles. This approach conveys two key types of information to the model. First, by highlighting these objects, the model is explicitly directed to analyze the visual characteristics of the pedestrian, including aspects such as gender, clothing, or location. Second, the precise placement of these bounding boxes enhances the VLM's ability to capture the spatial relationships between entities, such as their relative positions and orientations within the scene.

Furthermore, we leverage 3D gaze information to represent the pedestrian's attention by drawing a gaze line directly onto the image based on the given coordination. This visual representation allows us to understand where a pedestrian is looking within the scene and potentially determine the pedestrian's line of sight and awareness.

\subsubsection{Textual role-play prompting with hints}
Designing a detailed prompt plays a crucial role in enhancing the quality of generated descriptions, particularly in image and video captioning tasks. Recent research on prompting~\cite{kong2024betterzeroshotreasoningroleplay} has shown its effectiveness in improving model performance in zero-shot reasoning scenarios. 
%Moreover, \cite{sun2023autohintautomaticpromptoptimization} suggests that prompting with enriched instructions in multi-turn conversation can significantly improve the overall performance. 
Motivated by these findings, we propose a novel prompting framework augmented with attribute hints to guide the model toward producing more accurate and contextually appropriate descriptions.

% As shown in \ref{fig:analysispromptengineering}, the prompt lacked hints for spatial-invariant attributes may lead to incoherent captions, especially when the target captioning format in the dataset is well-structured and specific. To address this, we introduce structured hints that convey essential semantic patterns present in the desired outputs.

After carefully analyze the captions, we identified key attributes essential for high-quality spatial-invariant and temporal-variant descriptions. For the spatial-invariant captions, we incorporate attributes such as gender, age group, height, clothing details, surrounding context, and weather conditions. In the temporal-variant captions, we select hints that align with the focal actions and dynamics of the scene, ensuring that the model captures nuanced temporal-variant behaviors. 

% Specifically:
% \begin{itemize}
%     \item[-] \textbf{Hints for pedestrian caption}: body orientation, position, line of sight, visual state, fine-grained action, direction, speed and pedestrian awareness.
%     \item[-] \textbf{Hints for vehicle caption}: action and state, pedestrian visibility, relative position and speed.
% \end{itemize}

\subsection{Training Strategy}

Following caption decomposition, we fine-tune two separate Qwen2.5-VL-7B~\cite{bai2025qwen25vltechnicalreport} models using LoRA~\cite{hu2022lora}. Qwen2.5-VL-7B inherits strong capability in traffic analysis through its advanced spatial-invariant understanding and precise object localization. Each model is then tasked with specializing in processing image-text pairs corresponding to either spatial-invariant or temporal-variant components. The composition model, also fine-tuned using LoRA with structured and merged textual and visual prompts, refines the intermediate outputs into comprehensice captions that maintain the dense, structured format characteristic of our target domain. After caption tuning, we merge the LoRA weights with the base models to create stable checkpoints that incorporate the learned caption generation capabilities.

Building upon these checkpoints, we utilize them for the Visual Question Answering (VQA) task. This sequential adaptation strategy leverages the benefits of LoRA, which enables efficient initial specialization for caption generation. The two-stage approach ensures that models first establish robust captioning capabilities before adapting to downstream tasks, thereby minimizing interference between learning objectives and maintaining stable performance across both caption generation and VQA scenarios.

\section{Experiment}

\subsection{Dataset}

We utilized the WTS dataset~\cite{kong2024wts} and the BDD PC 5K, filtered pedestrian-related traffic scene videos based on BDD100K~\cite{BDD}  to fine-tune our model and validate the effectiveness of our method. The WTS dataset consists of 249 distinct scenarios with 810 videos from vehicles' dashboard and traffic surveillance cameras. The BDD dataset comprises 3402 pedestrian-centric videos from cars' dashboard cameras. All videos are divided into five segments. For each segment, two detailed descriptions surrounding the vehicle and pedestrian are provided, along with an average of four question-answer pairs about the current segment. Additionally, human-annotated and machine-generated bounding boxes for vehicles and pedestrians, as well as the 3D gaze of the person, are included as visual cues.

% A position of the pedestrian' head and direction of the person's 3d gaze are also provided in the WTS dataset.

\subsection{Metric and Implementation Details}
\subsubsection{Metric}
% To assess the quality of generated descriptions, we adopt four widely used language generation metrics: \textbf{BLEU-4}, \textbf{METEOR}, \textbf{ROUGE-L}, and \textbf{CIDEr}. BLEU-4 and ROUGE-L measure $n$-gram and sequence overlap with reference texts, METEOR accounts for linguistic features such as synonymy and word order, and CIDEr emphasizes consensus using TF-IDF weighted $n$-gram matching across multiple human annotations.

% For both internal and external evaluation subsets, we compute a \textbf{captioning score} as:

% % General normalized form
% \begin{equation}
%     C_{i/e} \;=\; \frac{\sum_{i=1}^{4} w_i \cdot M_i}{4}
% \end{equation}
% % \[
% % S \;=\; \sum_{i=1}^{4} w_i \cdot M_i
% % \]
% \textbf{where:}
% \begin{itemize}
%     \item w = [\,1,\,1,\,1,\,0.1\,], 
%     \item M = [\,\text{BLEU-4},\ \text{METEOR},\ \text{ROUGE-L},\ \text{CIDEr}\,]
% \end{itemize}

% The final captioning score is the average of internal and external captioning scores:
% \begin{equation}
% \text{Caption\_Score} = \frac{\text{C}_{\text{i}} + \text{C}_{\text{e}}}{2}
% \end{equation}

% Finally, the overall evaluation score combines the captioning performance with the VQA multiple-choice accuracy:
% \begin{equation}
% \text{S2} = \frac{\text{Caption\_Score} + \text{Acc}}{2} \times 100
% \end{equation}

We evaluate generated descriptions using four standard metrics: BLEU-4, METEOR, ROUGE-L, and CIDEr. BLEU-4 and ROUGE-L capture overlap with reference texts, METEOR accounts for linguistic similarity, and CIDEr emphasizes consensus across human annotations. The captioning score is a weighted average of these metrics, giving full weight to BLEU-4, METEOR, and ROUGE-L, and a lower weight to CIDEr. Scores are averaged across internal and external subsets. The final evaluation score combines the captioning score and VQA accuracy equally, scaled to a 100-point range.

\subsubsection{Implementation Details}
We use Qwen2.5-VL-72B and Qwen2.5-72B-Instruct~\cite{bai2025qwen25vltechnicalreport} for the reference caption enhancement and caption decomposition respectively. For task adaptation, the spatial-invariant and temporal-variant components are fine-tuned independently for $3$ epochs using LoRA~\cite{hu2022lora} with rank 64 and dropout rate $0.1$, and only the MLP layers are updated.  Training uses batch size 4 with 4-step gradient accumulation (effective batch size 16), AdamW optimizer with a learning rate of $2e^{-4}$, gradient checkpointing~\cite{feng2021optimal}, coupled with a cosine learning rate schedule and a  warm-up ratio of $10\%$. For the composition model, we increase gradient accumulation to 6 steps and reduce learning rate to 1e-4 for stable convergence. For VQA, we initialize from the best captioning checkpoints and fine-tune for one epoch with $lr=5e^{-5}$. All experiments use bf16 precision on a single NVIDIA RTX A6000 GPU.

\subsection{Ablation Study}

\begin{figure*}[htb]
    \centering
    \includegraphics[width=0.8\linewidth]{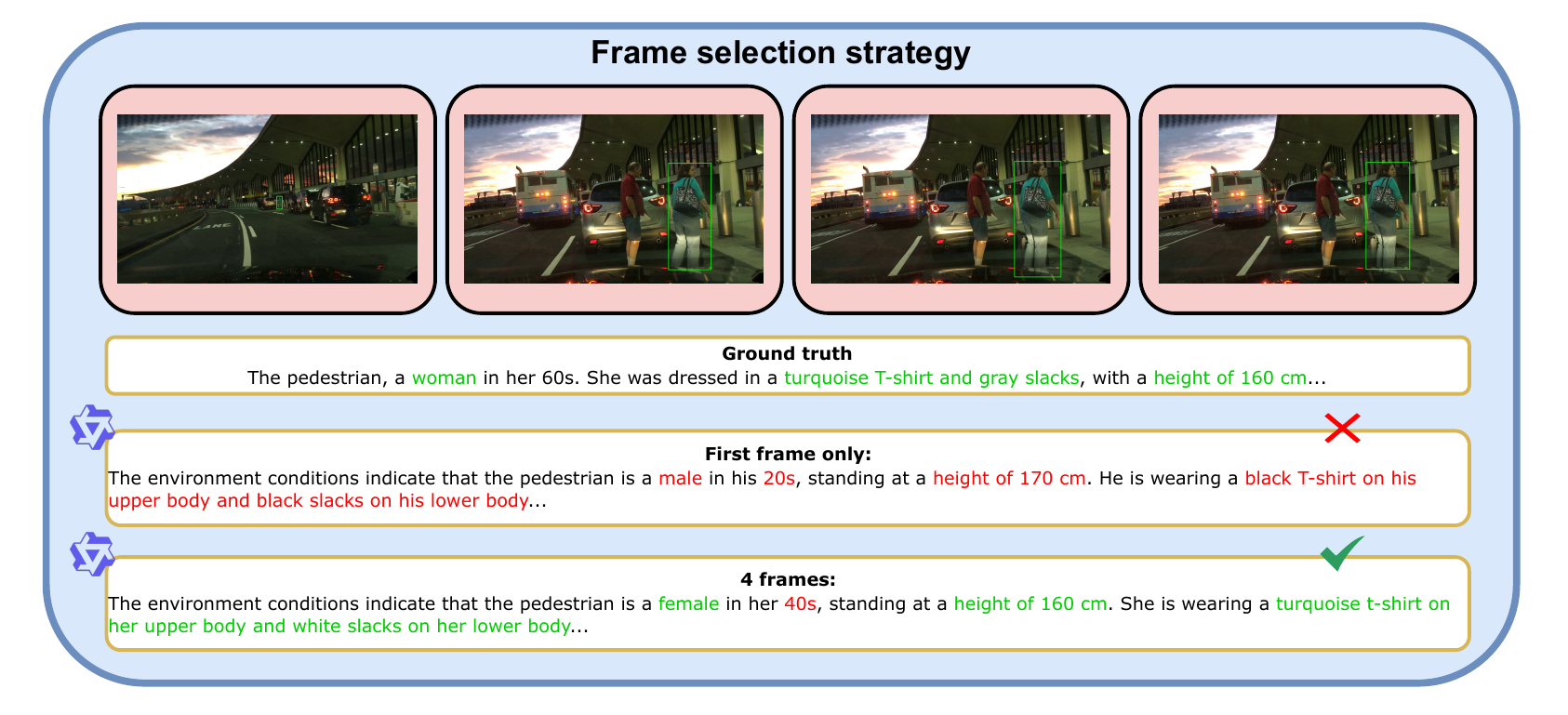} 
    \caption{Illustration of our frame selection strategy. The first two frames capture the pedestrian's appearance, while the last two provide environmental context. Using only a single first frame for inference leads to inaccurate descriptions due to limited visual evidence. In contrast, incorporating four frames results in more accurate and coherent captions. Temporal-variant components are excluded here for clearer visualization.}
    \label{fig:frame-selection-fig} 
\end{figure*}

\begin{table}[ht]
\centering
\small
\setlength{\tabcolsep}{4pt}
\begin{tabular}{lccccc}
\toprule
 & BLEU-4 & METEOR & ROUGE-L & CIDEr & \shortstack{Caption \\ Score} \\
\midrule
\multicolumn{6}{c}{\textbf{WTS}} \\
B & \textbf{0.230} & \textbf{0.431} & \textbf{0.422} & 0.822 & \textbf{29.160} \\
P & 0.216 & 0.424 & 0.410 & \textbf{0.938} & 28.595 \\
\midrule
\multicolumn{6}{c}{\textbf{BDD}} \\
B & 0.246 & 0.449 & 0.429 & 1.070 & 30.792 \\
P & \textbf{0.252} & \textbf{0.462} & \textbf{0.452} & \textbf{1.108} & \textbf{31.920} \\
\midrule
\multicolumn{6}{c}{\textbf{Combined}} \\
B & \textbf{0.238} & 0.440 & 0.426 & 0.946 & 29.975 \\
P & 0.234 & \textbf{0.443} & \textbf{0.431} & \textbf{1.023} & \textbf{30.258} \\
\bottomrule
\end{tabular}
\caption{Evaluation metrics (BLEU-4, METEOR, ROUGE-L, CIDEr, and Caption Score) for Baseline (B) and Pipeline (P) across WTS, BDD, and Combined subsets on the validation set.}
\label{tab:evaluation_metrics_baseline_pipeline}
\end{table}

\noindent \textbf{Effectiveness of Caption Decomposition Training} 
We present the potential of the caption decomposition approach in \textbf{Table~\ref{tab:evaluation_metrics_baseline_pipeline}}, comparing the baseline model, using only enhanced prompts, with the complete pipeline. Caption decomposition breaks down captions into semantic cues in traffic scenarios - such as appearance, environment, and location - allowing the model to focus on more specific, fine-grained details relevant in traffic scenarios. This leads to significant improvements in CIDEr by emphasizing semantic richness and rare traffic details. For instance, as shown in \textbf{Figure~\ref{fig:decomposition_caption}}, our caption "\textit{A male pedestrian in his 30s, approximately 170 cm tall. He was dressed in a white jacket, white hat, and black slacks.}" closely aligns with the diverse n-grams and specific content fragments used by human annotators "\textit{A male pedestrian in his 30s, approximately 170 cm tall, was observed [...] The pedestrian was dressed in a white jacket, white hat, and black slacks.}". Since CIDEr measures consensus by rewarding such rare but meaningful n-grams across multiple reference captions, our method effectively boosts its score. However, in the internal set, the decomposition disrupts exact phrase structures. Consequently, BLEU-4, METEOR, and ROUGE-L scores decline, reflecting a trade-off where semantic generalization reduces strict n-gram overlap under such perspectives. For example, our output might say "\textit{A male pedestrian in his 30s, approximately 170 cm tall.}" and "\textit{He was observed in a squatting position on a residential road.}" separately, instead of a reference like "\textit{A male pedestrian in his 30s, approximately 170 cm tall, was observed in a squatting position on a residential road.}" leading to mismatches in surface form despite capturing similar meaning. In contrast, the BDD 5K External set, comprised solely of vehicle-view videos with clearer pedestrian visibility and environmental context, uniformly benefits from caption decomposition. The more straightforward visual context allows the pipeline to generate captions that better capture salient scene elements and pedestrian behaviors, improving all metrics, and the overall caption score. 

Unlike our method, which partitions captions into spatial-invariant and temporal-variant components, the prior work~\cite{xuan2024divide} adopts a more fine-grained semantic decomposition strategy by categorizing caption fragments into five distinct types. This categorical framework allows for more flexible and semantically meaningful merging of segments, enabling systematic exploration of how different combinations affect performance. In contrast, our fixed structure constrains the space of possible merges, which can limit expressiveness and lead to suboptimal results. It is important to note that the relative decline in our performance is not attributable to the capabilities of the underlying model, but rather to the reduced flexibility inherent in our decomposition design.

\begin{figure*}[htb]
    \centering
    \includegraphics[width=0.9\linewidth]{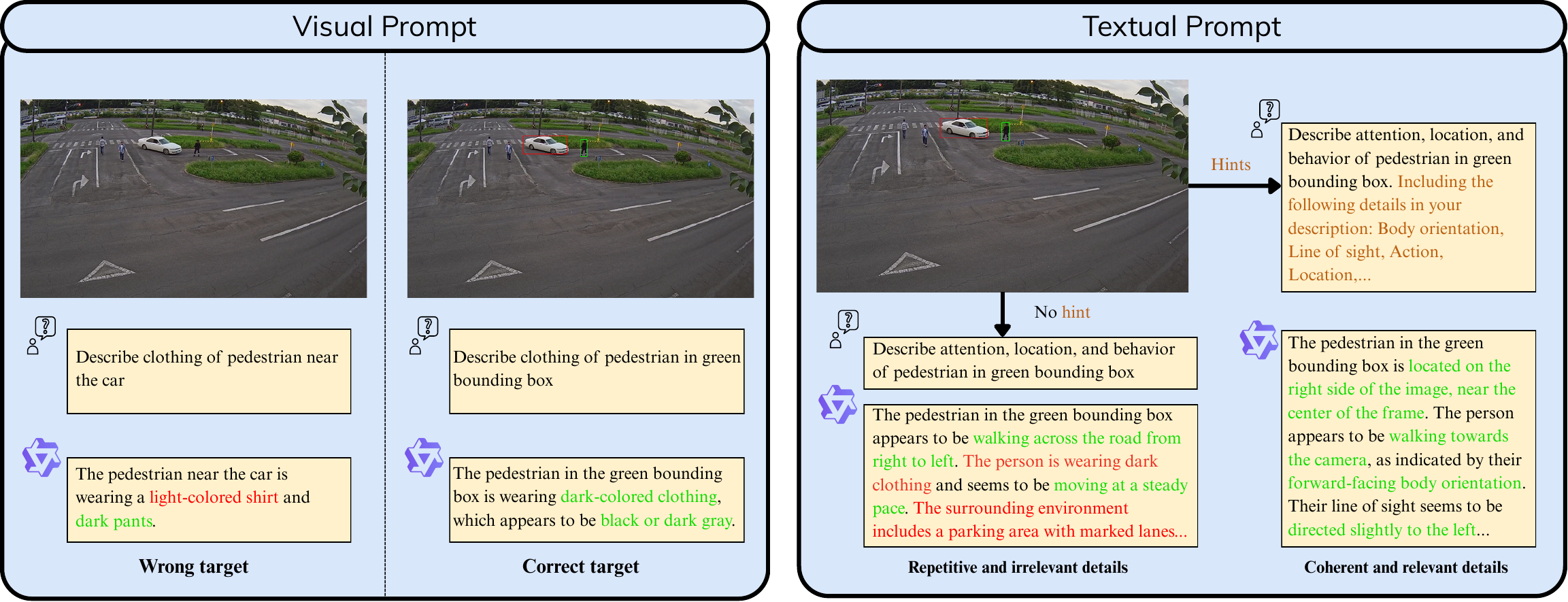} 
    \caption{Examples of visual prompts (Left) and corresponding textual prompts (Right).} 
    \label{fig:VPEaTPE} 
\end{figure*}

\noindent \textbf{Effectiveness of Frame Selection and Best-view filtering} In spatial-invariant parts, our frame selection strategy ensure enough frames for pedestrian's appearance and environmental context. As shown in \textbf{Figure~\ref{fig:frame-selection-fig}}, our frame selection strategy helps to capture details of the pedestrian's appearance, which the method use only first frame can not.
For the temporal-variant parts of the caption, the titles of each phase - prerecognition, recognition, judgment, action, and avoidance - reflect the general sequence of interactions between pedestrians and vehicles across distinct segments of the video. We posit that key behavioral transitions typically occur between the onset and conclusion of each phase, thereby allowing temporal dynamics to be effectively captured through phase-wise segmentation. 
The first frame typically marks the beginning of the interaction and, as demonstrated in prior work~\cite{CityLLaVA, xuan2024divide}, serves as an effective frame for captioning. The two additional frames selected from within the phase serve complementary purposes: the middle frame highlights the progression of the behavior, while the last frame reflects its resolution or end state. Collectively, these three frames offer a comprehensive overview of the full action span in phases 4 and 5, which demand fine-grained and temporally grounded descriptions of actions. 
% However, this sampling strategy may overlook important motion changes when critical actions occur between selected frames, for example in phases exceeding 9 seconds. This limitation arises from the uniform sampling approach, which may fail to capture short, sudden events. Moreover, as shown in Figure ~\ref{tab:frames-selection-fig}, there may be cases where the temporal parts can not be determined. Both inferred caption's and ground truth caption's position parts are contradicted to each other, which lead to noise in data. 
%However, we found a case where the temporal components cannot be fully described in a simple phrase. As illustrated on the right side of Figure~\ref{tab:frames-selection-fig}, two position part of ground truth are conflicted with each other, whether the pedestrian's position is based on first frame or third frame. % Sửa tiếp....

%To better address such abrupt changes, we encourage future research to explore adaptive frame sampling methods that can enhance model robustness to dynamic scene variations. 

\noindent \textbf{Effectiveness of Reference-Driven Understanding Approach} In our experiments, using references improved the performance of our captioning model, as shown in \textbf{Table~\ref{tab:reference-result-table}.}

\begin{table}[ht]
\centering
\small
\setlength{\tabcolsep}{2pt}
\begin{tabular}{cccccc}
\toprule
\shortstack[c]{Reference \\ -driven} & 
\shortstack[c]{BLEU-4} & 
\shortstack[c]{METEOR} & 
\shortstack[c]{ROUGE-L} & 
\shortstack[c]{CIDEr} & 
\shortstack[c]{Caption \\ Score} \\
% \shortstack[c]{\textbf{VQA} \\ \textbf{Score}} \\
\midrule
-- & 0.232 & 0.441 & 0.430 & 0.984 & 30.061 \\
\checkmark & \textbf{0.234} & \textbf{0.443} & \textbf{0.431} & \textbf{1.023} & \textbf{30.257} \\
\bottomrule
\end{tabular}
\caption{Captioning evaluation metrics for models without and with reference-driven guidance on the validation set of WTS and BDD.} 
\label{tab:reference-result-table}
\end{table}

% % Them bang ablation textual prompt
% \begin{table}[ht]
% \centering
% \small
% \begin{tabular}{ccc|c}
% \toprule
% \shortstack[c]{Appearance} & \shortstack[c]{Environment} & 
% \shortstack[c]{Caption Score} &
% \shortstack[c]{$\Delta$} \\
% \midrule
% \checkmark & \checkmark & 30.656 & - \\
% -- & \checkmark & 21.456 & -9.2 \\
% \checkmark & -- & 28.267 & -2.389 \\
% \bottomrule
% \end{tabular}
% \caption{Ablation studies (captioning validation score) of textual hints for spatial-invariant caption against inference with complete prompts.}
% \label{tab:fix_textual_prompt}
% \end{table}

\begin{figure*}[htb]
    \centering
    \includegraphics[width=1.0\linewidth]{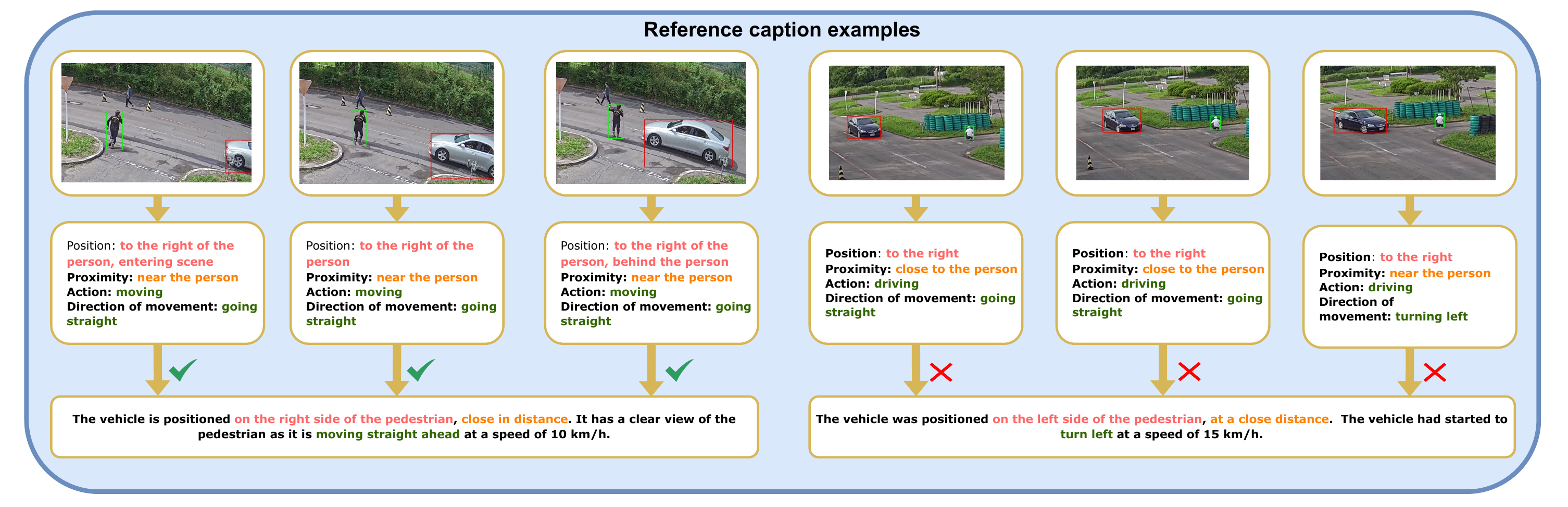} 
    \caption{Examples of captions with correct references (Left) and misinterpreted references (Right) from LVLM.} 
    \label{fig:Refcap-Figure} 
\end{figure*}

The improvement comes from the fact that captioning models require broad generalized knowledge and strong reasoning abilities to interpret complex traffic scenarios effectively. To support this, we use a large VLM to generate reference captions for each frame, serving as helpful hints rather than direct inputs to our final captions. This approach boosts the model’s confidence in most cases, particularly when recognizing difficult objects such as small, hidden, or blurry ones, or when reasoning about complex interactions. However, since a large VLM can sometimes misinterpret intricate details \textbf{Figure~\ref{fig:Refcap-Figure}}, we only use these references as guidance. This careful design allows us to benefit from the LVLM’s powerful visual and contextual understanding while avoiding the risk of inheriting incorrect information.

% \begin{table}[ht]
% \centering
% \small
% \setlength{\tabcolsep}{2pt}
% \begin{tabular}{cccccc}
% \toprule
% \shortstack[c]{Reference \\ -driven} & 
% \shortstack[c]{BLEU-4} & 
% \shortstack[c]{METEOR} & 
% \shortstack[c]{ROUGE-L} & 
% \shortstack[c]{CIDEr} & 
% \shortstack[c]{Caption \\ Score} \\
% % \shortstack[c]{\textbf{VQA} \\ \textbf{Score}} \\
% \midrule
% -- & 0.232 & 0.441 & 0.430 & 0.984 & 30.061 \\
% \checkmark & \textbf{0.234} & \textbf{0.443} & \textbf{0.431} & \textbf{1.023} & \textbf{30.257} \\
% \bottomrule
% \end{tabular}
% \caption{Captioning evaluation metrics for models without and with reference-driven guidance on the validation set of WTS and BDD.} 
% \label{tab:reference-result-table}
% \end{table}

\begin{table}[ht]
\centering
\begin{tabular}{ccc|cc}
\toprule
Visual & Textual & 
\shortstack{Spatial \\ Model} & 
\shortstack{Temporal \\ Model} \\
\midrule
\checkmark & --         & 17.113 & 25.245 \\
--     & \checkmark     & \textbf{31.847} & \underline{28.964} \\
\checkmark & \checkmark & \underline{30.656} & \textbf{29.185} \\
\bottomrule
\end{tabular}
\caption{Validation captioning scores on the valdation set for spatial and temporal models using visual prompts, textual prompts with hints, or both.}
\label{tab:prompt-ablation}
\end{table}

% Them bang ablation textual prompt
\begin{table}[ht]
\centering
\small
\begin{tabular}{ccc|c}
\toprule
\shortstack[c]{Appearance} & \shortstack[c]{Environment} & 
\shortstack[c]{Caption Score} &
\shortstack[c]{$\Delta$} \\
\midrule
\checkmark & \checkmark & 30.656 & - \\
-- & \checkmark & 21.456 & -9.2 \\
\checkmark & -- & 28.267 & -2.389 \\
\bottomrule
\end{tabular}
\caption{Ablation studies (captioning validation score) of textual hints for spatial-invariant caption against inference with complete prompts.}
\label{tab:fix_textual_prompt}
\end{table}

% Bảng so sánh score prompt: không visual, không textual, có cả 2
\noindent \textbf{Effectiveness of Instruction Optimization} As shown in \textbf{Table~\ref{tab:prompt-ablation}}, combining textual hints with visual prompts yields the best performance in the temporal model, achieving a score of 29.185 - outperforms when using visual prompts only by 3.94. In contrast, the spatial model benefits most from textual hints alone, outperforming the visual prompt only by 14.734 and the combined prompt by 1.191. The slight performance drop of the combined prompt in the spatial model is caused by missing or incomplete bounding box annotations, which in some cases fail to cover the pedestrian, leading to misleading visual cues for the model. As shown in \textbf{Figure~\ref{fig:VPEaTPE}}, the proposed paradigm utilizes textual and visual prompts to enhance the accuracy of fine-grained perception and to precisely focus on target objects and relevant regions.% Chen image 
We also investigate the impact of textual role-play prompting with carefully designed hints by conducting two ablation experiments targeting spatial-invariant and temporal-variant captioning tasks separately. For each task, we systematically remove certain groups of textual hints as shown in \textbf{Figure~\ref{fig:intro}} to examine how different types of hints affect captioning scores.  
In the spatial-invariant descriptions, removing either the Appearance or Environment group of hints leads to noticeable drops in performance, as reported in \textbf{Table~\ref{tab:fix_textual_prompt}}. The largest drop, 9.2 points, occurs when Appearance hints are removed, suggesting their critical role. For example, without Appearance hints, descriptive terms related to color become overly general and less informative, as shown in \textbf{Figure~\ref{fig:VPEaTPE}}. In contrast, the removal of Environment hints results in a smaller decrease of 2.389 points, indicating that although they contribute to spatial-invariant understanding, their impact is less substantial than that of Appearance hints.
For the temporal-variant descriptions, removing the Action, Attention, or Location group of hints leads to consistent decreases in performance, with the largest drop of 1.29 points occurring when Action hints are removed, as shown in \textbf{Table~\ref{tab:not_fix_textual_prompt}}. This suggests that each group of temporal-variant hints contributes to supporting temporal coherence and the overall descriptive quality of the generated captions.  
These findings underscore the importance of textual role-play prompting with structured hints in guiding both spatial-invariant and temporal-variant comprehension in captioning tasks.

% \begin{table}[ht]
% \centering
% \begin{tabular}{ccc|cc}
% \toprule
% Visual & Textual & 
% \shortstack{Spatial \\ Model} & 
% \shortstack{Temporal \\ Model} \\
% \midrule
% \checkmark & --         & 17.113 & 25.245 \\
% --     & \checkmark     & \textbf{31.847} & \underline{28.964} \\
% \checkmark & \checkmark & \underline{30.656} & \textbf{29.185} \\
% \bottomrule
% \end{tabular}
% \caption{Validation captioning scores on the valdation set for spatial and temporal models using visual prompts, textual prompts with hints, or both.}
% \label{tab:prompt-ablation}
% \end{table}

% % Them bang ablation textual prompt
% \begin{table}[ht]
% \centering
% \small
% \begin{tabular}{ccc|c}
% \toprule
% \shortstack[c]{Appearance} & \shortstack[c]{Environment} & 
% \shortstack[c]{Caption Score} &
% \shortstack[c]{$\Delta$} \\
% \midrule
% \checkmark & \checkmark & 30.656 & - \\
% -- & \checkmark & 21.456 & -9.2 \\
% \checkmark & -- & 28.267 & -2.389 \\
% \bottomrule
% \end{tabular}
% \caption{Ablation studies (captioning validation score) of textual hints for spatial caption against inference with complete prompts.}
% \label{tab:fix_textual_prompt}
% \end{table}

\begin{table}[ht]
\centering
\small
\begin{tabular}{cccc|c}
\toprule
\shortstack[c]{Action} & 
\shortstack[c]{Attention} & 
\shortstack[c]{Location} & 
\shortstack[c]{Caption Score} &
\shortstack[c]{$\Delta$} \\
\midrule
\checkmark & \checkmark & \checkmark & 29.185 & - \\
-- & \checkmark & \checkmark & 27.895 & -1.29 \\
\checkmark & -- & \checkmark & 28.277 & -0.908 \\
\checkmark & \checkmark & -- & 28.157 & -1.028 \\
\bottomrule
\end{tabular}
\caption{Ablation studies (captioning validation score) of textual hints for temporal-variant caption against inference with complete prompts.}
\label{tab:not_fix_textual_prompt}
\end{table}

\begin{table}[ht]
\centering
\small
\begin{tabular}{cc}
\toprule
Finetuning method & VQA \\
\midrule
Direct finetune & 75.367 \\
2-stage & \textbf{81.507} \\
\bottomrule
\end{tabular}
\caption{Impact of two-stage training on VQA scores on the official test data.}
\label{tab:optimal-learning}
\end{table}

\begin{table}[ht]
\centering
\begin{tabular}{lc}
\toprule
Team Name & Score \\
\midrule
CHTTLIOT & 60.039 \\
SCU\_Anastasiu & 59.118 \\
Metropolis\_Video\_Intelligence & 58.848 \\
ARV & 57.913 \\
Rutgers ECE MM & 57.465 \\
VNPT\_AI & 57.113 \\
\textbf{AIO\_GENAI4E} & \textbf{55.655} \\
Tyche & 52.148 \\
\bottomrule
\end{tabular}
\caption{Leaderboard scores of different teams on test set.}
\label{tab:leaderboard_scores}
\end{table}

% \noindent \textbf{Effectiveness of Caption Decomposition Training} 
% We present the potential of the caption decomposition approach in Table \ref{tab:evaluation_metrics_baseline_pipeline}, comparing the baseline model, using only enhanced prompts, with the complete pipeline. Caption decomposition breaks down captions into semantic cues in traffic scenarios. This leads to significant improvements in CIDEr by emphasizing semantic richness and rare traffic details. However, in the internal set, which consists of multi-view videos with domain-specific phrasing, decomposition disrupts exact phrase structures due to the caption parser model . Consequently, BLEU-4, METEOR, and ROUGE-L scores decline, reflecting a trade-off where semantic generalization reduces strict n-gram overlap under such perspectives.
% In contrast, the BDD 5K External subset, comprised solely of vehicle-view videos with clearer pedestrian visibility and environmental context, uniformly benefits from caption decomposition. The more straightforward visual context allows the pipeline to generate captions that better capture salient scene elements and pedestrian behaviors, improving all metrics, and the overall caption score. These results indicate that decomposition enhances semantic generalization and caption quality, particularly in vehicle-centric views, while also revealing viewpoint-dependent trade-offs between semantic capture and phrase-level matching. 

\noindent \textbf{Effectiveness of 2-stage Training Strategy for VQA} In \textbf{Table~\ref{tab:optimal-learning}} demonstrates that our two-stage LoRA-based fine-tuning approach, where the model is initially fine-tuned for structured caption generation, enabling it to develop a deeper understanding of visual content, significantly enhances its subsequent fine-tuning for the VQA (Visual Question Answering) task. This sequential process results in a more accurate and stable model, yielding a 6.14\% improvement in VQA accuracy.

\section{Conclusion}
This paper presents a computationally efficient approach for enhancing vision-language models in traffic safety analysis through \textbf{caption decomposition}, \textbf{intelligent frame selection}, and \textbf{reference-driven understanding}. Our methodology successfully balances performance with computational efficiency, achieving a competitive score of 55.655 and ranking 7th among participating teams on the evaluation leaderboard \textbf{Table~\ref{tab:leaderboard_scores}}. The experimental results validate the effectiveness of each proposed component, demonstrating that with resource efficiency, we can still achieve leaderboard performance.

{
    \small
    \bibliographystyle{ieeenat_fullname}
    \bibliography{main}
}


\end{document}